\documentclass{article}

\usepackage[preprint,nonatbib]{neurips_2025}

\usepackage[utf8]{inputenc} 
\usepackage[T1]{fontenc}    
\usepackage{hyperref}       
\usepackage{url}            
\usepackage{booktabs}       
\usepackage{amsfonts}       
\usepackage{nicefrac}       
\usepackage{microtype}      
\usepackage{xcolor}         

\usepackage{graphicx}

\usepackage{amsmath}
\usepackage{amssymb}
\usepackage{mathtools}
\usepackage{amsthm}
\usepackage[capitalize,noabbrev]{cleveref}

\usepackage{color,soul}
\usepackage{wrapfig}
\usepackage{graphicx}
\usepackage{multirow}
\usepackage{caption}

\usepackage{multicol}
\usepackage{graphicx}
\usepackage{textcomp}
\usepackage{times}
\usepackage{epsfig}
\usepackage{lipsum}
\usepackage{tabularx}

\usepackage{algorithmic}
\usepackage{newfloat}
\usepackage{listings}

\usepackage{multirow}
\usepackage{graphicx}
\usepackage{caption}
\usepackage{xspace}
\usepackage{gensymb}
\usepackage{siunitx} 
\usepackage{graphicx}
\usepackage{array}
\usepackage{colortbl}

\theoremstyle{plain}

\theoremstyle{definition}

\theoremstyle{remark}

\usepackage{xcolor}

\definecolor{yijiecolor}{RGB}{0,0,255}       

\usepackage{amsmath,amsfonts,bm}

\def\eqref#1{equation~\ref{#1}}

\def\1{\bm{1}}

\DeclareMathAlphabet{\mathsfit}{\encodingdefault}{\sfdefault}{m}{sl}
\SetMathAlphabet{\mathsfit}{bold}{\encodingdefault}{\sfdefault}{bx}{n}

\title{Hierarchical Scoring with 3D Gaussian Splatting for Instance Image-Goal Navigation}

\author{%
  Yijie Deng\thanks{Equal contribution.}\enspace$^{1,2,3,4}$, Shuaihang Yuan$^\ast$$^{1,2,4}$, Geeta Chandra Raju Bethala$^{1,2,4}$, \AND Anthony Tzes$^{1,2}$, Yu-Shen Liu$^{5}$, Yi Fang\thanks{Corresponding author: Yi Fang <yfang@nyu.edu>.}\enspace$^{1,2,3,4}$\\
  \\
  $^{1}$NYUAD Center for Artificial Intelligence and Robotics (CAIR), Abu Dhabi, UAE.\\
  $^{2}$New York University Abu Dhabi, Electrical Engineering, Abu Dhabi 129188, UAE.\\
  $^{3}$New York University, Electrical \& Computer Engineering Dept., Brooklyn, NY 11201, USA.\\
  $^{4}$Embodied AI and Robotics (AIR) Lab, NYU Abu Dhabi, UAE.\\
  $^{5}$School of Software, Tsinghua University, Beijing, China. \\
}

\begin{document}
\maketitle

\begin{abstract}
Instance Image-Goal Navigation (IIN) requires autonomous agents to identify and navigate to a target object or location depicted in a reference image captured from any viewpoint. While recent methods leverage powerful novel view synthesis (NVS) techniques, such as three-dimensional Gaussian splatting (3DGS), they typically rely on randomly sampling multiple viewpoints or trajectories to ensure comprehensive coverage of discriminative visual cues. This approach, however, creates significant redundancy through overlapping image samples and lacks principled view selection, substantially increasing both rendering and comparison overhead. In this paper, we introduce a novel IIN framework with a hierarchical scoring paradigm that estimates optimal viewpoints for target matching. Our approach integrates cross-level semantic scoring, utilizing CLIP-derived relevancy fields to identify regions with high semantic similarity to the target object class, with fine-grained local geometric scoring that performs precise pose estimation within promising regions. Extensive evaluations demonstrate that our method achieves state-of-the-art performance on simulated IIN benchmarks and real-world applicability.

\end{abstract}    
\section{Introduction}

\label{sec:intro}
Instance Image-Goal Navigation (IIN) is critical in embodied navigation, requiring an agent to identify and move to the object or location depicted in a target image—often captured from any viewpoint \cite{krantz2022instance}. This flexibility is essential in real-world scenarios where users may provide photos from arbitrary perspectives. However, viewpoint discrepancies, cluttered scenes, and occlusions complicate the alignment of target images with the agent's observations. Effective solutions must robustly align these visual representations, enabling the agent to accurately interpret and navigate to the specified object or location.

Motivated by advances in novel view synthesis (NVS) methods, such as Neural Radiance Fields (NeRF) \cite{mildenhall2021nerf} and three-dimensional Gaussian splatting (3DGS) \cite{kerbl20233d}, recent approaches have begun to explore more expressive, view-consistent scene representations for IIN. Methods \cite{cui2024frontier,wang2024lookahead} combine NeRF rendering with a topological graph, embedding RGB observations and learned image features into graph nodes. While this strategy retains more detailed appearance information, discretizing the environment into nodes constrains the agent's ability to observe scenes from diverse angles or navigate more complex layouts, thereby limiting truly free-view navigation. 

Alternatively, 3DGS-based approaches \cite{lei2025gaussnav, meng2024beings, honda2025gsplatvnm} preserve a continuous three-dimensional representation, offering high geometric fidelity and robust performance. However, these methods typically rely on randomly sampling multiple viewpoints \cite{lei2025gaussnav} or trajectories \cite{meng2024beings, honda2025gsplatvnm} to ensure comprehensive coverage of discriminative visual cues. This sampling strategy in continuous 3D space creates significant redundancy through overlapping rendered images, substantially increasing both rendering and comparison overhead. The resulting trade-off between coverage and efficiency limits the practical deployment of these approaches.

To address these limitations, we introduce a novel IIN framework with a hierarchical scoring paradigm that efficiently estimates optimal viewpoints for target matching. Our approach eliminates excessive sampling by integrating two complementary scoring mechanisms over the 3DGS environment. First, our global semantic scoring leverages CLIP-derived relevancy fields to identify regions with high semantic similarity to the target object class. By computing cosine similarity between CLIP text embeddings of the detected object class and features of each Gaussian, followed by thresholding and diffusion, we form coherent candidate regions containing potential target objects. Second, our local geometric scoring employs a two-stage approach: initially performing region-level scoring by comparing sampled rays from candidate regions with the goal image's DINOv2 features through cross-attention, then conducting precise pose estimation within the most promising region.

\begin{figure}[t]
    \centering
    \includegraphics[width=\linewidth]{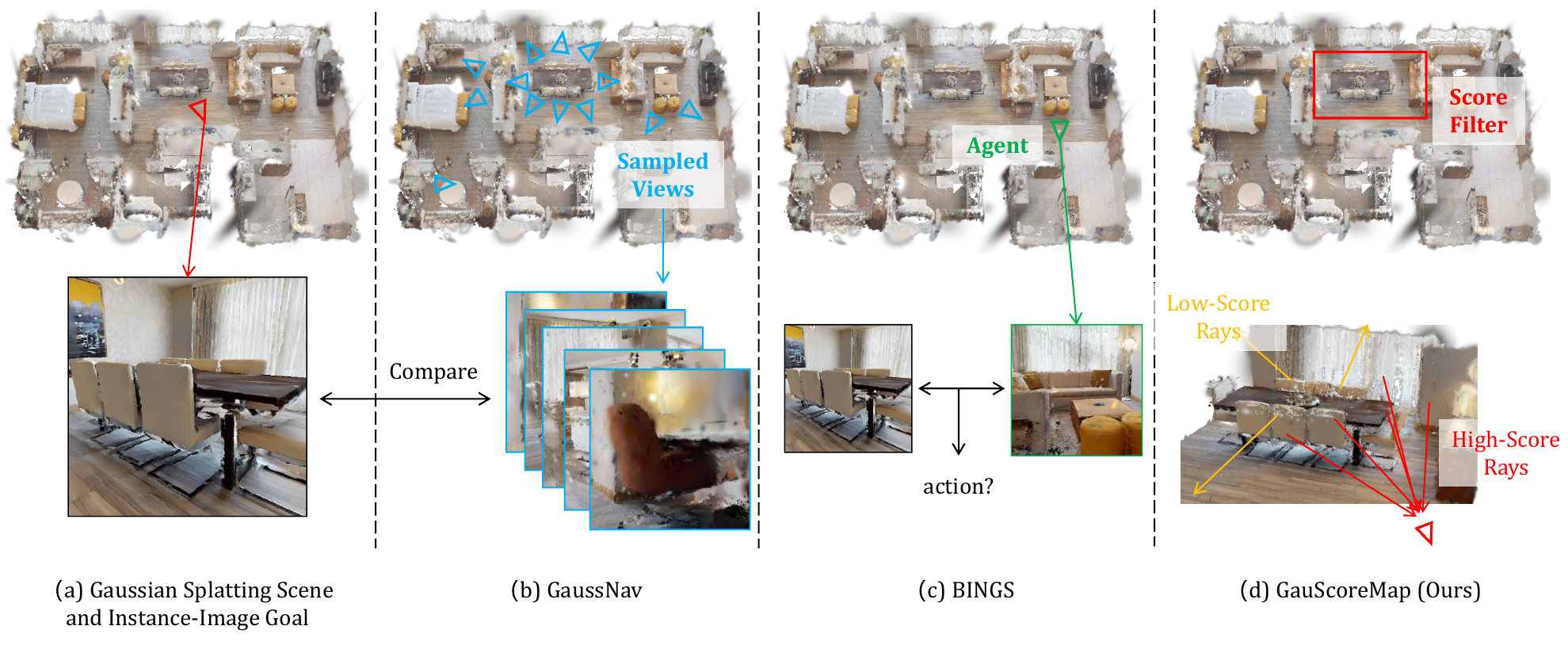}
    \caption{Overall method comparison between our method and two gaussian splatting-based navigation methods.}
    \label{fig:comparison}
\end{figure}

Figure \ref{fig:comparison} illustrates the key differences between our method and existing Gaussian splatting-based approaches. GaussNav \cite{lei2025gaussnav} exhaustively samples views around predicted object instances without strategic guidance, resulting in inefficient matching. BEINGS \cite{meng2024beings} relies on a learned policy to measure differences between observations and the target image, but struggles when these images share minimal similarity. In contrast, our method requires only two forward passes to generate global semantic and local geometric score maps that effectively localize the target object. The target pose is then determined by sampling Gaussian-ray pairs once and using triangulation among high-score rays, significantly reducing computational requirements while maintaining accuracy.
We empirically validate the effectiveness and efficiency of this approach, demonstrating state-of-the-art performance on the simulated benchmark and real-world applicability.

The contributions of our method are mainly summarized as follows: 1) We introduce a two-tier scoring approach that combines high-level semantic alignment with fine-grained geometric matching, producing a continuous relevance map that highlights where the target image content is most likely to appear in the 3D environment. 2) We leverage the score map to identify and select the most informative viewpoints for matching, thereby obviating the need for exhaustive or random sampling throughout the environment. 3) We achieve new state-of-the-art results on instance-specific image-goal navigation benchmark data and further demonstrate our method's reliable operation in real-world indoor environments.

\section{Related Work}
\label{sec:rw}

\subsection{Instance Image Goal Navigation}
\label{sec:rw_IGN}
Deep reinforcement learning has emerged as one major approach to IIN, where policies are learned end-to-end to align an agent’s current observations with a target reference image. Methods in this category often demonstrate promising results in simulator settings by leveraging extensive training data to handle viewpoint changes and partial observations \cite{lei2024instance,qin2025navigatediff}. However, these reactive strategies frequently struggle to retain knowledge of previously visited areas in cluttered or large-scale scenes \cite{krantz2023navigating}, as they lack an explicit representation of the environment’s structure. Consequently, performance typically degrades when the agent encounters complex layouts or needs to re-localize after losing sight of salient features. 

To address the need for context retention and better adaptability, a second line of IIN research incorporates environment mapping to guide navigation \cite{yu2023l3mvn,majumdar2022zson,yuangamap}. By representing space with either metric or topological maps, these systems maintain a record of observed areas, helping the agent efficiently revisit or avoid previously explored regions. Early map-based methods focused on metric representations—often in the form of 2D bird’s-eye view grids derived from SLAM pipelines—that allow the robot to infer its spatial location relative to obstacles and landmarks \cite{chaplot2020neural,lei2024instance}. While this approach mitigates some shortcomings of purely reactive strategies, the 2D format discards much of the 3D geometry and fails to preserve high-fidelity texture details essential for matching a goal image. More recently, structured maps and novel view synthesis techniques \cite{cui2024frontier,wang2024lookahead,lei2025gaussnav,meng2024beings,honda2025gsplatvnm} have been explored for navigation, combining visual features with graph nodes to better retain rich appearance cues. These latter developments underscore an emerging consensus that robust IIN demands a representation capable of bridging environment structure with the high-resolution visual information vital to pinpointing the target location.

\subsection{Novel View Synthesis in Embodied Visual Navigation}
\label{sec:rw_nvs}
Early efforts such as e2e-NeRF-nav \cite{liu2024integrating} integrate an online Neural Radiance Field directly into the control loop, enabling end-to-end training of both memory and policy. However, continual NeRF \cite{mildenhall2021nerf} updates make each step computationally expensive. Building on the idea that a radiance field can serve as more than passive memory, HNR-VLN \cite{wang2024lookahead} uses NeRF \cite{mildenhall2021nerf} to render candidate future viewpoints so a graph search can anticipate rewards before acting, shifting complexity from policy learning to look-ahead synthesis. Frontier-enhanced Topological Memory \cite{cui2024frontier} improves upon this approach by stitching “ghost” nodes onto a classical topological graph, combining geometric reachability with appearance-based reasoning at exploration frontiers. While this approach maintains detailed appearance information, representing the environment as discrete nodes constrains the agent’s capacity to observe scenes from diverse viewpoints and may hinder navigation in more complex layouts. Instead of combining graph-based structures with NeRF, GaussNav \cite{lei2025gaussnav} employs 3D Gaussian splitting \cite{kerbl20233d} to represent the environment, preserving high-fidelity geometry and high-frequency textures. BEINGS \cite{meng2024beings} compares global descriptors of the target image and the current observation, shifting all computation to Monte Carlo model-predictive control and rendering hypothetical rollouts each time step. While effective, 3DGS based methods suffers from high computational overhead. These observations motivate the development of a method that leverages fine-grained local visual information without requiring extensive trajectory or viewpoint sampling.
\section{Method}
\label{sec:method}

\subsection{Overview}

\begin{figure}[t]
    \centering
    \includegraphics[width=\linewidth]{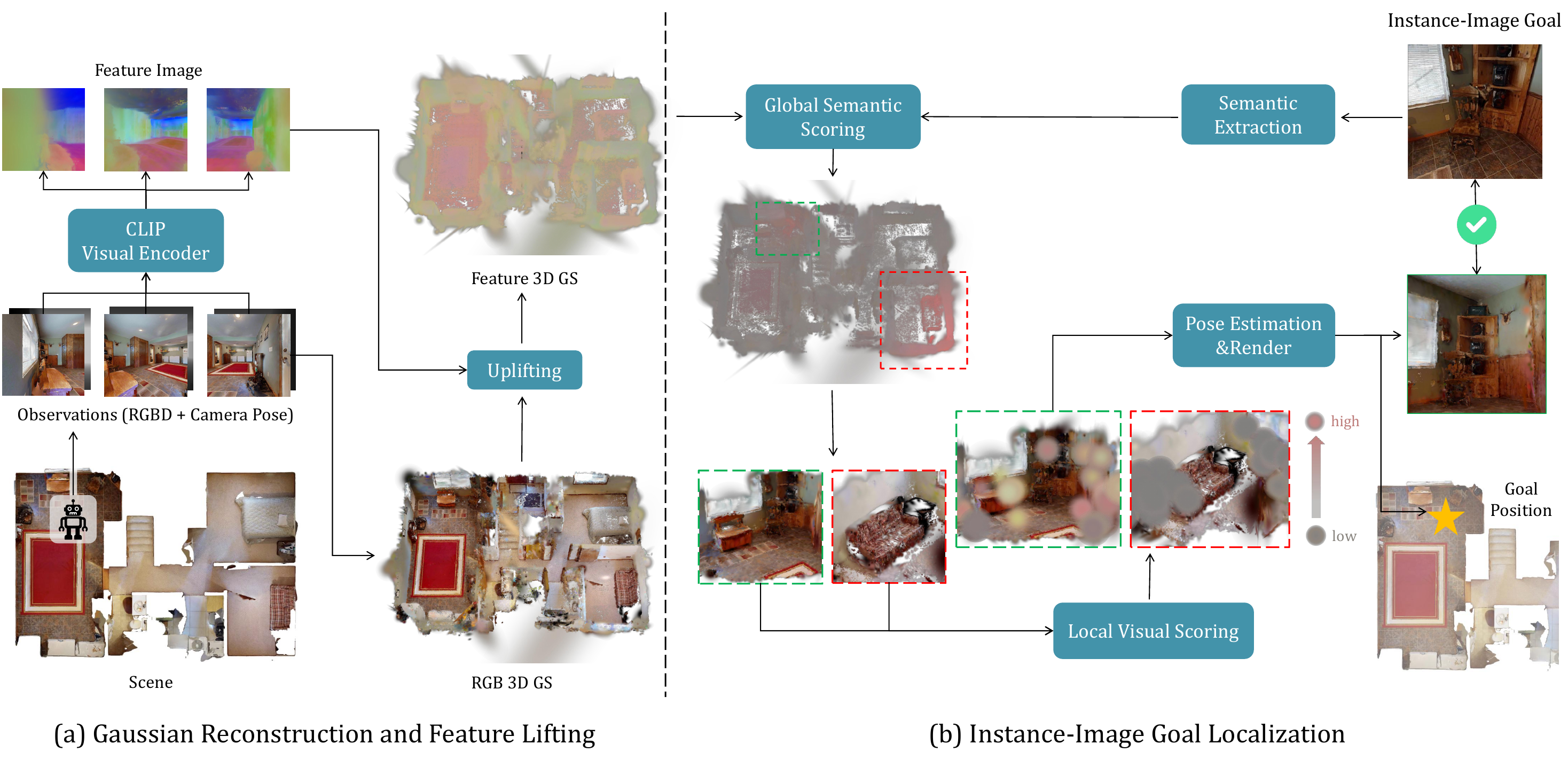}
    \caption{Overview of our GauScoreMap approach for Instance Image-goal Navigation. Our method consists of two main stages: (a) Gaussian Reconstruction and Feature Lifting, where we build a 3D Gaussian representation of the environment and lift CLIP features into this representation; and (b) Instance-Image Goal Localization, which uses a two-step scoring process to first identify semantically relevant regions and then precisely locate the target object instance.}
    \label{fig:pipeline}
\end{figure}

In Instance Image-goal Navigation (IIN), an agent navigates to a specific object instance shown in a goal image $I_g$. Starting from a random position $p_0$ and orientation $r_0$, the agent receives RGB images $I_t$, depth maps $D_t$, and camera poses $P_t$ at each timestep. Based on these observations, it selects actions $a_t$ to locate the target. An episode is considered successful if the agent reaches the vicinity of the goal and can observe the target object by adjusting its camera orientation within a maximum action limit.

To address this IIN task, we introduce a modular approach called \textit{Gaussian Splatting Score Maps for Visual Navigation} (GauScoreMap), illustrated in Figure \ref{fig:pipeline}. Our method consists of two main stages: Gaussian Reconstruction and Feature Lifting, and Instance-Image Goal Localization. When deployed in a new environment, the agent first explores the scene to gather observations and constructs an RGB Gaussian splatting representation. This representation serves as a foundation for lifting 2D visual features into 3D space, creating a comprehensive feature-rich Gaussian field. Once constructed, we proceed to the second stage, which comprises two sequential scoring processes. First, for a given instance image, we extract its semantic information to compute relevancy scores against the feature field, generating a global semantic score map that identifies candidate regions of interest. Then, we compute a more precise local score map by measuring the similarity between the goal image and these pre-filtered regions within the Gaussian field, enabling accurate pose estimation and precise localization of the target object instance. Finally, the agent plans an efficient path to the target using the identified location.

\subsection{Gaussian Reconstruction and Feature Lifting}

\subsubsection{Gaussian Reconstruction}

When placed in a new environment, the agent employs a frontier-based exploration strategy~\cite{yamauchi1997frontier,holz2010evaluating,julia2012comparison} to systematically cover the environment and collect observations for Gaussian reconstruction.

From the collected observations $\{(I_i,D_i,P_i)|i\in[0,N]\}$ (RGB images, depth maps, and camera poses), we reconstruct the RGB Gaussian splatting field using a hierarchical approach~\cite{yugay2023gaussian,yugay2024magic}. We first build sub-Gaussian maps of local regions before merging them into a global representation. For each subset of observations $\{(I_i,D_i,P_i)|i\in[m,n]\}$, we initialize a submap from the first frame by backprojecting the RGB image into 3D space using depth and pose information. As we process subsequent frames, the submap is densified with additional Gaussian primitives, and finally, these local submaps are merged into a comprehensive global field.

During training, we render both color and depth from the Gaussian field:
\begin{equation}
\hat{I}_i(u,v) = \sum_{j=1}^{K} c_j \alpha_j \prod_{k=1}^{j-1}(1-\alpha_k), \quad
\hat{D}_i(u,v) = \sum_{j=1}^{K} d_j \alpha_j \prod_{k=1}^{j-1}(1-\alpha_k)
\end{equation}
where $\hat{I}_i(u,v)$ and $\hat{D}_i(u,v)$ are rendered color and depth values, $c_j$ and $d_j$ are color and depth values of the $j$-th Gaussian along the ray through pixel $(u,v)$, $\alpha_j$ is the opacity, and $K$ is the number of intersected Gaussians. The optimization uses a combined loss:
\begin{equation}
\mathcal{L} = \lambda_{\text{color}} \mathcal{L}_{\text{color}} + \lambda_{\text{depth}} \mathcal{L}_{\text{depth}} + \lambda_{reg}\mathcal{L}_{reg}
\end{equation}
where $\mathcal{L}_{\text{color}} = \|I_i - \hat{I}_i\|_1$ and $\mathcal{L}_{\text{depth}} = \|D_i - \hat{D}_i\|_1$ are L1 losses between ground truth and rendered images/depths, $\mathcal{L}_{reg}$ regularizes Gaussian scaling, and $\lambda$ terms balance the loss components.

\subsubsection{Feature Lifting}


The visual features produced by the CLIP\cite{radford2021learning} visual encoder are uplifted with simple aggregation from all collected frames~\cite{marrie2024ludvig}. For each 3D Gaussian in the scene, we construct its feature representation as a weighted average of 2D features from all frames. Specifically, a 2D feature $F_{i,(u,v)}$ from the $i$-th image at pixel coordinate $(u,v)$ contributes to the feature $f_g$ of a Gaussian $g$ by a factor proportional to the rendering weight $w_g(i,(u,v))$, if the Gaussian $g$ belongs to the ordered set $S_{i,(u,v)}$ associated with that image-pixel pair. Denoting $S_g = \{(i,(u,v)) | g \in S_{i,(u,v)}\}$ as the set of image-pixel pairs contributing to the feature $f_g$, the resulting features are defined as:

\begin{equation}
f_g = \sum_{(i,(u,v))\in S_g} \bar{w}_g(i,(u,v))F_{i,(u,v)} \text{ with } \bar{w}_g(i,(u,v)) = \frac{w_g(i,(u,v))}{\sum_{(i,(u,v))\in S_g} w_g(i,(u,v))}
\end{equation}


\subsection{Global Semantic Scoring}
\label{method:global}

After constructing the CLIP feature Gaussian field, we leverage it to generate a global relevancy score map based on the visual input. Since an image contains information more than the target instance object, directly computing the relevancy between the goal image with the feature field might result in noisy segmentation. Therefore, we use a strong instance object detector such as Mask-RCNN\cite{he2017mask} to first extract the class label as the text input to the clip text encoder, then we compute its CLIP text embedding $E_T$. Then, for each Gaussian $g$ in our field with CLIP feature $f_g$, we calculate a relevancy score $S_g$ using cosine similarity:

\begin{equation}
S_g = \frac{E_T \cdot f_g}{\|E_T\| \cdot \|f_g\|}
\end{equation}


This operation assigns a relevancy score to each Gaussian in our 3D representation, creating a continuous score field throughout the environment. By applying a threshold $\tau$ to this score field, we obtain segmented regions of Gaussians likely belonging to instances of the target category. However, naive thresholding often results in fragmented and noisy segments. To address this issue, we follow LUDVIG~\cite{marrie2024ludvig} by incorporating scene geometry and diffusing the segmentation to larger, more coherent regions based on feature similarity between neighboring Gaussians. This diffusion process connects fragmented parts of the same object instance, resulting in well-formed connected components. Each of these connected components then serves as a candidate region containing a potential target object. By identifying these candidate regions, we significantly reduce the search space for the subsequent stage. These well-defined local contexts enable more efficient and accurate image-based 6D pose estimation, where we match the goal image against each candidate region to precisely locate the specific target instance.


\subsection{Local Geometric Scoring}

After identifying multiple candidate regions through global semantic scoring, we need to further refine our search in two stages: first determining the most likely local region containing the target object, and then estimating the precise 6D pose within that region.

\subsubsection{Local scoring for region selection}

For each candidate region identified in the global scoring stage, we sample random Gaussians and generate rays in the hemisphere defined by the surface normal of each Gaussian (estimated using neighboring Gaussians). This process yields a set of ray inputs: $\{(o_i, d_i, c_i)|i\in[0,K]\}$, where $o_i$ is the ray origin, $d_i$ is the ray direction, and $c_i$ is the 1-st order spherical harmonic coefficient representing the color of the ray.

Following the approach in~\cite{matteo20246dgs}, we encode these rays using a learned MLP with positional encoding:

\begin{equation}
r_i = \text{MLP}(\gamma(o_i), \gamma(d_i), \gamma(c_i))
\end{equation}

where $\gamma(\cdot)$ denotes the positional encoding function. This transforms the ray set into a feature representation of shape $(K, C_1)$.

Concurrently, we process the goal image $I_g$ through a DINOv2~\cite{oquab2023dinov2} visual encoder to get its visual feature $F_g$ of shape $(l, C_2)$, where $l = h \times w$ represents the spatial dimensions of the feature map.

These features are then compared through a cross-attention mechanism:

\begin{equation}
A = \text{CrossAttention}(r, F_g) \in \mathbb{R}^{K \times l}
\end{equation}

By summing along the second dimension, we obtain a region-level score map $S_r \in \mathbb{R}^K$, where $S_{r_i} = \sum_{j=1}^{l} A_{i,j}$. The region with the highest aggregate score is selected as the most likely location of the target object.

\subsubsection{Fine-Grained Pose Estimation for Precise Localization}

Once we've identified the most promising region, we perform a second, more dense sampling of Gaussian-ray pairs within this region. This denser sampling allows for more precise localization of the target object. 
We select top $k$ Gaussian-ray pairs with the highest scores from the new samples, and perform triangulation as described in~\cite{matteo20246dgs} to estimate the 6D pose (position and orientation) of the target object. This two-stage scoring approach—first at the region level and then at the pose level—enables our system to efficiently narrow down the search space before performing precise localization, significantly improving both the efficiency and accuracy of the object localization process.
These two scoring steps use the same pretrained ray-image cross attention neural network by minimizing the difference between the predicted camera 6D pose and the gt camera 6D pose as ~\cite{matteo20246dgs}.

\subsection{Path Planning}

The local geometric scoring stage yields a precise target position for the agent to navigate towards. For safe navigation through the environment, we project each Gaussian splat as a 3D point onto a 2D Bird's Eye View (BEV) map, creating a grid representation that distinguishes between navigable areas and obstacles. With this comprehensive global 2D map, the agent computes the optimal trajectory from its current position to the target location using the Fast Marching Method~\cite{sethian1999fast}, ensuring efficient and collision-free navigation.




\section{Experiment}
\label{sec:exp}

\begin{table}
  \caption{Success rate and SPL comparison of our method with four sets of baseline methods.}
  \label{tb:main}
  \centering
  \begin{tabular}{llll}
    \toprule
    Category & Method & SR $\uparrow$ & SPL $\uparrow$ \\
    \midrule
    \multirow{4}{*}{End-to-end}
    &RL Baseline~\cite{krantz2022instance}          & 0.083   & 0.035  \\
    &OVRL-v2 ImageNav~\cite{yadav2023habitat}     & 0.006   & 0.002  \\
    &OVRL-v2 IIN~\cite{yadav2023habitat}          & 0.248   & 0.118  \\
    &FGPrompt~\cite{sun2023fgprompt}             & 0.099   & 0.028  \\
    \midrule
    \multirow{3}{*}{MultiON Transfer}
    &MultiON Baseline~\cite{wani2020multion}     & 0.066   & 0.045  \\
    &MultiON Implicit~\cite{marza2023multi}     & 0.143   & 0.107  \\
    &MultiON Camera~\cite{chen2022learning}       & 0.186   & 0.142  \\
    \midrule
    \multirow{3}{*}{SOTA IIN}
    &Mod-IIN~\cite{krantz2023navigating}              & 0.561   & 0.233  \\
    &IEVE Mask RCNN~\cite{lei2024instance}       & 0.684   & 0.241  \\
    &IEVE InternImage~\cite{lei2024instance}     & 0.702   & 0.252  \\
    \midrule
    \multirow{5}{*}{\parbox{3cm}{SOTA IIN with Scene/3DGS Map}}
    &Mod-IIN (Scene Map)~\cite{krantz2023navigating}             & 0.563   & 0.323  \\
    &IEVE Mask RCNN (Scene Map)~\cite{lei2024instance}      & 0.683   & 0.331  \\
    &IEVE InternImage (Scene Map)~\cite{lei2024instance}    & 0.705   & 0.347  \\
    &GaussNav (3DGS Map)~\cite{lei2025gaussnav}             & 0.725   & 0.578  \\
    &GauScoreMap (3DGS Map)          & \textbf{0.784}   & \textbf{0.605}  \\
    \bottomrule
  \end{tabular}
\end{table}


\subsection{Experiment Setup}

\textbf{Dataset.} We conduct our experiments using the Habitat~\cite{szot2021habitat} simulator. For scene data, we utilize the Habitat-Matterport 3D dataset (HM3D)~\cite{yadav2023habitat}, which comprises over 200 scenes of 3D reconstructions of real-world indoor environments with semantic annotations. These scenes are divided into 145, 36, and 35 scenes for training, validation, and testing, respectively. Specifically, we use version 0.2 of the HM3D dataset and follow the Instance ImageGoal Navigation (IIN)~\cite{krantz2022instance} in the Habitat Navigation Challenge 2023\footnote{https://aihabitat.org/challenge/2023/}. We train our local geometric scoring function on the HM3D training set and fintune on the new environment using the collected observations. Then we evaluate our method on the 1,000 validation episodes specified by Krantz \textit{et al.}~\cite{krantz2022instance}. This validation subset encompasses six object categories: \{\textit{chair, couch, bed, toilet, television}\} and includes 795 unique object instances.

\textbf{Agent Configuration.} We adopt the standard agent configuration from the Habitat Navigation Challenge 2023. The agent is modeled as a rigid-body cylinder with zero turning radius, standing 1.41m tall with a radius of 0.17m. A forward-facing RGB-D camera is mounted at a height of 1.31m. At each time step $t$, the agent receives observations consisting of RGB images, depth maps, and sensor poses. The agent operates in a continuous action space with four dimensions: \textit{linear velocity, angular velocity, camera pitch velocity, and velocity stop}. Each action dimension accepts values between -1 and 1, which are then scaled according to their respective configuration parameters. The maximum linear speed is $35cm/frame$, while the maximum angular velocity is $60^{\circ}/frame$.

\textbf{Evaluation Metrics.} Our evaluation incorporates both effectiveness and navigation efficiency metrics. The primary metrics we use are SR (Success Rate) and SPL (Success weighted by Path Length). A navigation attempt is considered successful when the agent executes the stop action within a 1.0m radius of the target object and can visually detect the object by adjusting its camera orientation. The SPL metric, as introduced by Anderson et al.~\cite{anderson2018evaluation}, provides a balanced assessment of navigation efficiency by considering both success and path optimality. 

\subsection{Comparison with State-of-the-art Methods}

We compare our method against a comprehensive set of baseline approaches as presented in Table \ref{tb:main}, with baseline results sourced from GaussNav~\cite{lei2025gaussnav}. The comparison methods are organized into four categories:

\textbf{End-to-end Baselines.} These methods learn navigation policies directly from raw observations: (1) RL Baseline, which processes RGB, depth, goal image, and positional information through separate encoders before combining them with an LSTM and training with PPO; (2) OVRL-v2 ImageNav~\cite{yadav2023habitat}, which employs self-supervised pretraining for visual encoders but performs poorly (0.006 SR) without fine-tuning; (3) OVRL-v2 IIN, a fine-tuned version specifically for the IIN task, achieving significantly better results (0.248 SR); and (4) FGPrompt~\cite{sun2023fgprompt}, which leverages fine-grained goal prompting for more effective navigation.

\textbf{MultiON Transfer Methods.} These approaches were originally designed for the MultiON task, which shares similarities with scene-specific map representations: (1) MultiON Baseline, a standard implementation of ~\cite{wani2020multion}; (2) MultiON Implicit~\cite{marza2023multi}, which learns an implicit neural representation; and (3) MultiON Camera~\cite{chen2022learning}, which develops an active camera movement policy. These methods directly receive the semantic category of the target object as input.

\textbf{State-of-the-art IIN Methods.} Leading IIN approaches include: (1) Mod-IIN~\cite{krantz2022instance}, which decomposes the task into exploration, goal instance re-identification, goal localization, and local navigation; (2) IEVE Mask RCNN~\cite{lei2024instance}, which implements a modular architecture using Mask RCNN~\cite{he2017mask} for object detection; and (3) IEVE InternImage, an enhanced variant with a more powerful detector.

\textbf{SOTA IIN with Scene/3DGS Map.} This category includes the above methods when augmented with different map representations: (1-3) Mod-IIN, IEVE Mask RCNN, and IEVE InternImage with traditional scene maps; (4) GaussNav~\cite{lei2025gaussnav}, which utilizes 3D Gaussian Splatting maps; and (5) our proposed approach, which also leverages 3DGS maps but enhances performance.

As shown in Table~\ref{tb:main}, our method significantly outperforms all baselines, achieving the highest success rate (0.784) and SPL (0.605). Notably, our approach surpasses GaussNav by 5.9\% in success rate and 2.7\% in SPL. While both GaussNav~\cite{lei2025gaussnav} and our method utilize Gaussian splatting fields for navigation, our superior performance stems from enhanced localization capabilities through two key improvements.



This substantial improvement demonstrates that our hierarchical scoring paradigm—combining cross-level semantic scoring with fine-grained geometric scoring—effectively leverages the rich texture details preserved in the 3DGS representation. By first identifying candidate regions through CLIP-derived relevancy fields and then performing local geometric scoring and precise 6D pose estimation, our approach enables more accurate target localization without requiring exhaustive viewpoint sampling or additional verification steps. We further illustrate the real-world applicability of our approach in a video demonstration, which can be found in the supplementary materials.

\begin{table}
  \caption{Ablation study of our method.}
  \label{tb:ablation}
  \centering
  \begin{tabular}{lll}
    \toprule
    Method & SR $\uparrow$ & SPL $\uparrow$ \\
    \midrule
    GauScoreMap & 0.784 & 0.605 \\
    GauScoreMap w.o. Global Semantic Scoring & 0.608 & 0.419 \\
    GauScoreMap w.o. Local Geometric Scoring & 0.421 & 0.310 \\
    GauScoreMap w.o. Finetuning & 0.721 & 0.543 \\
    GauScoreMap w. GT Match & 0.842 & 0.650 \\
    GauScoreMap w. GT Global Localization & 0.944 & 0.742 \\
    \bottomrule
  \end{tabular}
\end{table}

\subsection{Ablation Study}

We ablate the design choices of our method and show their inflences on the final performance in Table ~\ref{tb:ablation}. And we analyze each module of our method:

\textbf{GauScoreMap w.o. Global Semantic Scoring.} The global semantic scoring module serves as a prefilter to extract candidate local regions for finer local localization. From Table \ref{tb:ablation}, we can see that without global semantic scoring, with only local geometric scoring to produce the predicted target position, the success rate drops by 17.6\%. This is because indoor scenes have severe occlusion and complicated spatial distributions. Simply sampling Gaussians and computing the relationships between ray features and Gaussians suffers from ambiguity and inefficiency. The global semantic scoring effectively narrows down the search space by identifying semantically relevant regions first, allowing the local geometric scoring to focus on promising areas. This two-stage approach significantly improves both accuracy and computational efficiency compared to relying solely on local visual scoring.

\textbf{GauScoreMap w.o. Local Geometric Scoring.} When we remove the local geometric scoring component and rely only on global semantic scoring, performance decreases dramatically with a 36.3\% drop in success rate. This substantial decline highlights the critical role of fine-grained geometric matching in precisely localizing the target object. While global semantic scoring can identify candidate regions containing objects of the target category, it lacks the precision to distinguish specific object instances with similar semantic properties. The local geometric scoring module provides this crucial instance-level discrimination by establishing detailed geometric correspondences between the goal image features and the 3D scene representation.

\textbf{GauScoreMap w.o. Finetuning.} Our local geometric scoring module benefits from finetuning its network parameters when deployed in a new scene. This adaptation is efficiently implemented by utilizing the observations already collected during the Gaussian reconstruction phase, allowing the network to adjust to the specific visual characteristics of each environment. Without this finetuning step, we observe a 6.3\% decrease in performance. This demonstrates the value of scene-specific adaptation, as indoor environments exhibit significant variation in visual appearance, lighting conditions, and texture details that can affect feature matching accuracy.

\begin{figure}[t]
  \centering
  \includegraphics[width=\linewidth]{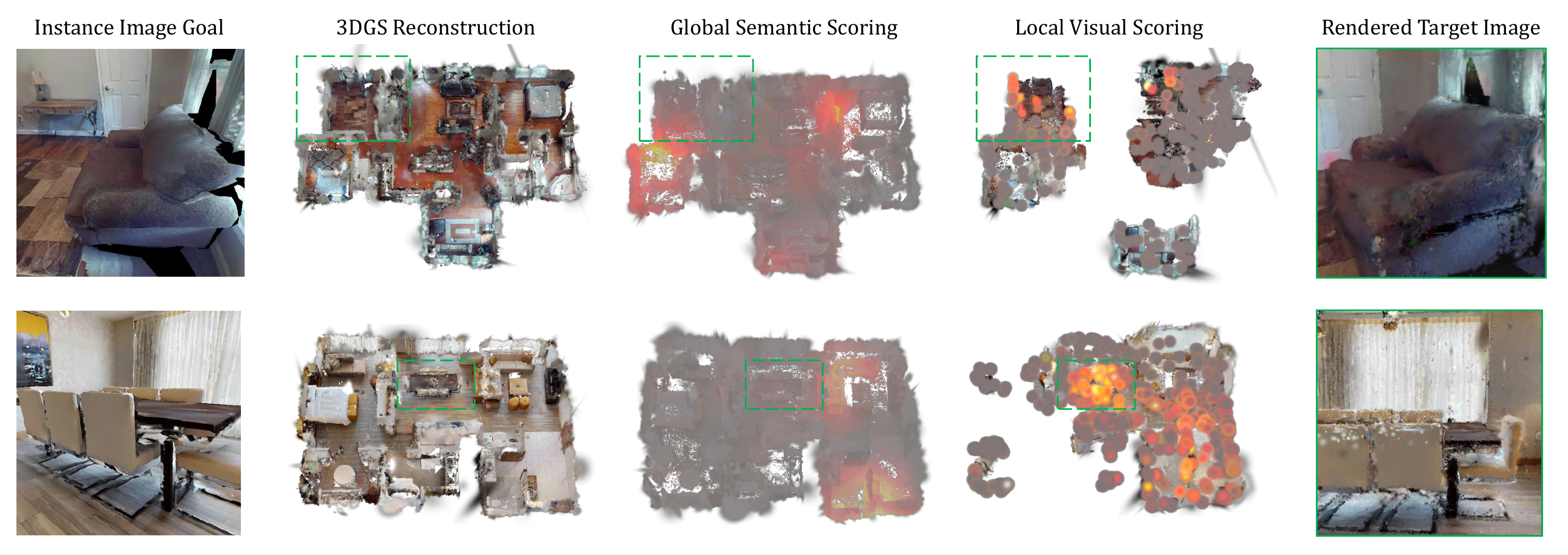}
  \caption{Complete scoring and localization processes of two scenes in HM3D~\cite{yadav2023habitat} validation set. }
  \label{fig:extra_gaussian}
\end{figure}

\begin{figure}[t]
  \centering
  \includegraphics[width=\linewidth]{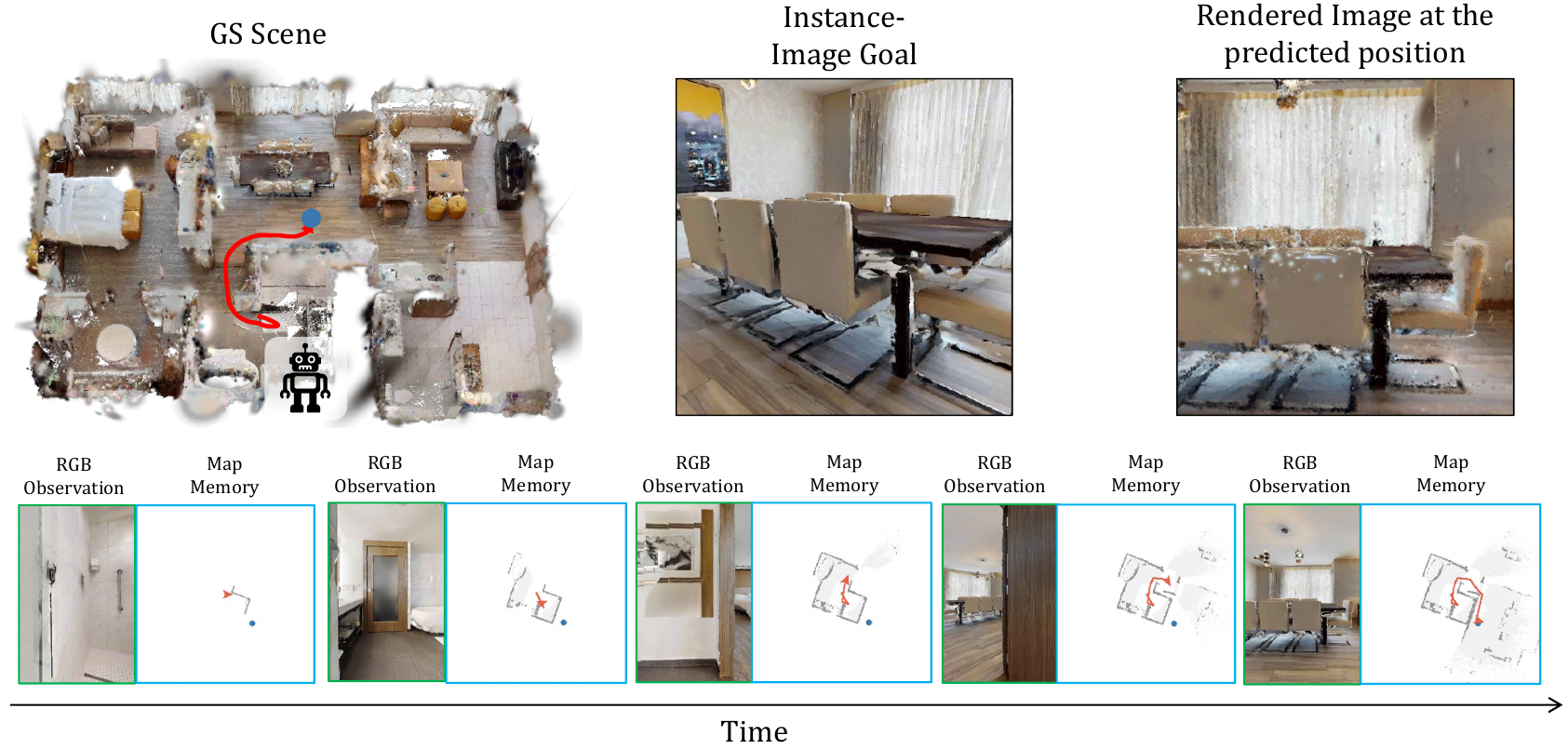}
  \caption{The process of the agent navigating to the position of the instance image goal.}
  \label{fig:navigation}
\end{figure}

\subsection{Additional Visual Results}

Figure~\ref{fig:extra_gaussian} shows two instance image goal localization examples from the HM3D~\cite{yadav2023habitat} validation set. These demonstrate our two-stage approach: global semantic scoring first identifies relevant regions, then local geometric scoring refines this by generating concentrated scores that precisely pinpoint the target. Figure~\ref{fig:navigation} shows a complete navigation sequence to a target. The visualization shows how the agent uses the identified target location (blue dot on the map) to plan and execute an efficient path, finally adjusting its orientation to view the target from the correct perspective.



\section{Conclusion}
\label{sec:concl}
In this work, we introduce a novel Instance Image-Goal Navigation framework that directly tackles the principal challenges of viewpoint variation, semantic ambiguity, and complex scene layouts. By combining cross-level semantic scoring with fine-grained geometric scoring, the method yields a continuous score map that obviates the need for exhaustive or random viewpoint sampling.
Empirical evaluations on simulated benchmarks confirm state-of-the-art performance, underscoring the method’s effectiveness and practical applicability. Furthermore, we deploy the proposed approach on a humanoid agent and validate its performance in real-world indoor environments. A key limitation of our method is that it focuses primarily on static environments and does not account for dynamic scenes with moving objects or changing layouts. Future directions may include augmenting this framework with large language model (LLM) reasoning.

\bibliographystyle{unsrt}
\bibliography{ref}

\end{document}